# Bone marrow cells detection: A technique for the microscopic image analysis

Haichao Cao[1*], Hong Liu[2], Enmin Song[3]

[123] Huazhong University of Science and Technology, Hong Shan District, Wuhan, Hubei, China
[*] dr_caohaichao@163.com

**Abstract**: In the detection of myeloproliferative, the number of cells in each type of bone marrow cells (BMC) is an important parameter for the evaluation. In this study, we propose a new counting method, which also consists of three modules including localization, segmentation and classification. The localization of BMC is achieved from a color transformation enhanced BMC sample image and stepwise averaging method (SAM). In the nucleus segmentation, both SAM and Otsu's method will be applied to obtain a weighted threshold for segmenting the patch into nucleus and non-nucleus. In the cytoplasm segmentation, a color weakening transformation, an improved region growing method and the K-Means algorithm are used. The connected cells with BMC will be separated by the marker-controlled watershed algorithm. The features will be extracted for the classification after the segmentation. In this study, the BMC are classified using the SVM, Random Forest, Artificial Neural Networks, Adaboost and Bayesian Networks into five classes including one outlier, namely, neutrophilic split granulocyte, neutrophilic stab granulocyte, metarubricyte, mature lymphocytes and the outlier (all other cells not listed). Our experimental results show that the best average recognition rate is 87.49% for the SVM.

## 1. Introduction

Bone marrow cells (BMC) detection is an important approach for the discovery and diagnosis of leukaemia and other blood-related diseases. Currently, there are many studies about for segmentation and detection of white blood cells in peripheral blood, but few studies on BMC. Due to the fat and dense distribution in the bone marrow, it is difficult for the pathological detection. Traditional BMC detection methods require that the human eye carefully observe the bone marrow samples under the microscope classifies and counts the BMC manually. These manual detection methods are not only time-consuming, but also easily lead to human errors. With the development of image processing technology, some advanced detection techniques have been used for the diagnosis of microscopic images. However, an accurate detection of BMC is still a very difficult task due to some factors such as different staining methods and lighting condition, which can make a variety of microscopic images. To deal with the complexity of BMC images, we propose a detection strategy which consists on a combination of colour transformation, region growing [1], K-Means clustering [2], watershed algorithm [3] and machine learning [4].

In the past decade, many solutions have been proposed for segmenting white blood cells in peripheral blood. In [5], the authors proposed a novel segmentation scheme which detects well on regular cells. However, it cannot be applied to the larger connected cells. In [6], the authors proposed a clustering method based on rough set for leukocyte segmentation. In [7], the authors proposed the stepwise merging rules and gradient vector flow snake method for leukocyte segmentation. It works well under the condition of normal staining. However, the processing time is very slow. In [8], a computer-aided segmentation method which is based on Otsu's method [9] for segmenting leukocyte was proposed. In [10], the authors proposed the Yager's fuzzy method which processes three channel images of R, G and B to obtain three segmented images. These three-segmented images are merged to obtain the final segmentation result. In [11], the authors proposed an automatic segmentation method based on the intuitionistic fuzzy divergence based thresholding which is robust even in the influence of noise. However, it performs poor for the low staining and weak edge information. In [12], the authors used Atanassov's intuitionistic fuzzy sets [13, 14] and interval Type II fuzzy sets theory to carry out the segmentation of blood leukocyte images. Compared with other algorithms, it is more effective when used on the segmentation of an image containing more than one cell. In [15], the authors adopted an unsupervised segmentation of leukocytes images using thresholding neighborhood valley-emphasis and deconvolution techniques to separate methylene blue and eosin in Giemsa stained images for processing. In [16], the authors developed a new leukocyte segmentation method based on intuitionistic fuzzy divergence to achieve the robust segmentation performance. In [17], the authors proposed a novel method for leukocyte image segmentation, which is based on feed forward neural network with random weights. In [3], the authors proposed a new algorithm to segment normal cells and leukaemia cells in peripheral blood and bone marrow. The algorithm models the color and shape characteristics of white blood cells by two transformations, which are used in the marker-controlled watershed algorithm. In [18], the authors proposed a method based on Gram-Schmidt Orthogonalization along with a snake algorithm to segment nucleus and cytoplasm of the cells.

Of course, some solutions have also been proposed for classifying white blood cells in peripheral blood. The method proposed in [19] for an automatic leucocyte recognition is based on a fuzzy divergence technique. In [20], the authors integrate Fisher linear discriminant pre-processing with feedforward neural networks for classifying cultured cells in the bright field. In [21], the authors proposed a new detection algorithm based on fuzzy cellular neural network which combines the advantages of threshold segmentation followed by mathematical morphology and fuzzy logic method. In [22], Boosting-based method is used for automatic detection of leukocytes in blood smear images. In [23], the authors proposed a white blood cell automatic counting system based on mathematical models, principal component analysis, and neural networks. In [24], naïve Bayes classifiers are employed to sort the cells into eosinophil, lymphocyte, monocyte and neutrophil. In [25], the authors derive more precise regional boundaries by using the gradient-based region growing method and classify the regions using the fuzzy and non-fuzzy techniques based on the features of shape, size, color, and texture. In [26], the authors proposed a computer-aided diagnostic method for the diagnosis of myeloma cells.

Our contributions in this work are summarized as follows:
1) To the best of our knowledge, we first proposed a strategy for various types of white blood cells detection in bone marrow images.
2) We propose a novel thresholding method and two color transformation methods.
3) We propose to integrate thresholding method, color transformation, region growing, K-Means clustering, watershed algorithm, and machine learning for cell detection.

The organization of the paper is structured as follows: Section 2 outlines the theoretical concepts of BMC localization, segmentation, feature extraction, and classification. Section 3 gives our experimental results and discussion. The conclusion then follows.

## 2. Materials and methods

This paper presents a new method for the detection of BMC. It consists of localization, segmentation, feature extraction and classification. We only consider five types of BMC in our detection, namely, neutrophilic split granulocyte, neutrophilic stab granulocyte, metarubricyte, mature lymphocytes and the outlier (cells not listed). The flow chart of the proposed method is shown in Fig.1.

### 2.1. Localization of BMC

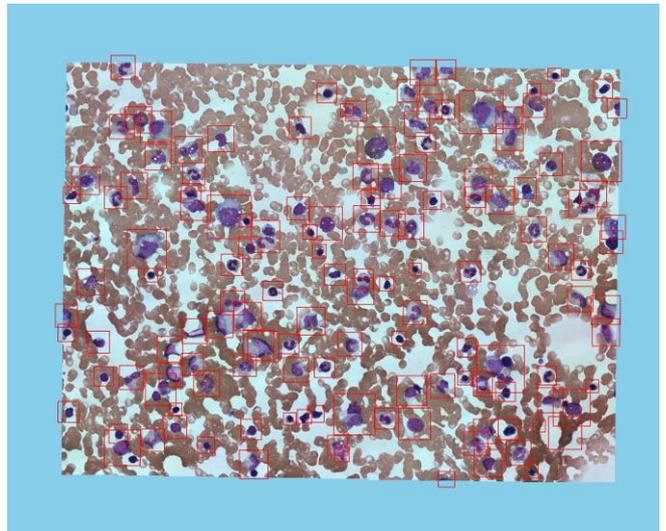

**Fig. 2.** *Localization of BMC in bone marrow samples*

#### 2.1.1 Color transformation

In general, after Wright's staining [27, 28] of the BMC, the color of nucleus is dark purple, which is darker than that of cytoplasm and mature red blood cells. The BMC image is usually stored in the RGB (Red-Green-Blue) format. In order to enhance the contrast between the nucleus of BMC and other cell tissues, it is necessary to convert from RGB to Hue-Saturation-Intensity (HSI). Our experiments show that the gray value of BMC nucleus is lower than non-nucleus pixels in the green channel, and the gray value distribution of BMC nucleus is the highest compared with non-nucleus pixels in the Hue (H) component. Similarly, it is also true for the Saturation (S) component. According to this observation, we can derive a color transformation for the BMC as shown in Eq. (1).

In Eq. (1), $I_H(x, y)$, $I_S(x, y)$ and $I_G(x, y)$ represent the gray values of a pixel in the coordinate (x, y) in the H channel, S channel and G channel, respectively. Parameters $w_1$, $w_2$, $w_3$ are real values in [0, 1] that can be adjusted to experimental results. In this study, the value of $w_1$, $w_2$, $w_3$ is 0.4, 0.6, and 1.0.

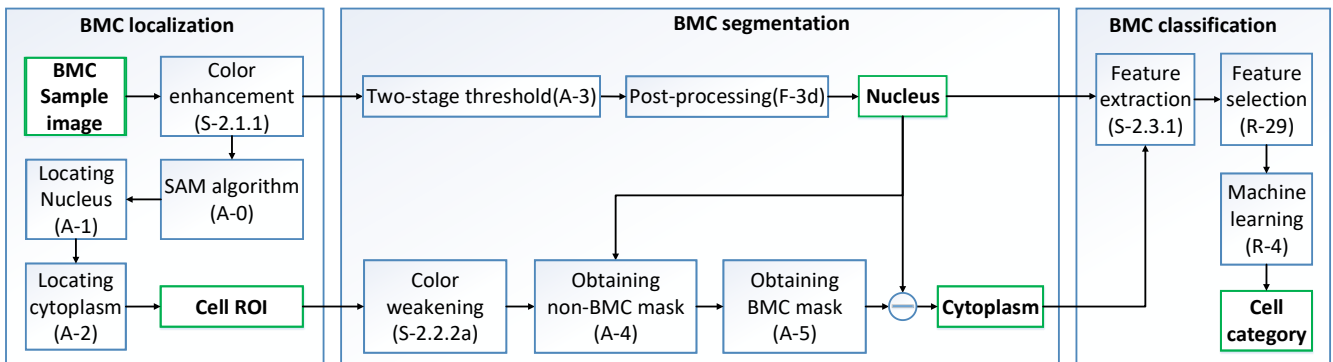

**Fig. 1.** *Flow chart of proposed BMC detection method. Note that, "A-n" denotes algorithm n; "S-c" denotes section c; "R-p" denotes references p; "F-i" denotes figure i.*

As shown in Fig.2, BMC samples usually contain a large number of mature red blood cells and impurities. How to accurately isolate BMC from bone marrow samples is a problem in the classification of BMC. To identify the location of BMC, we design a color transformation, stepwise averaging method and two-stage locating operation.

$$I_{HSG}(x, y) = \begin{cases} \dfrac{\omega_1 * I_H(x, y) + \omega_2 * I_S(x, y)}{\omega_3 * I_G(x, y)}, & I_G(x, y) > 0 \\ 255, & I_G(x, y) = 0 \end{cases} \quad (1)$$

The image obtained by Eq. (1) is called the Hue, Saturation, and Green color enhancement image (HSG). In contrast to the image in each of H, S, and G channels, the color enhancement image significantly highlights the region of BMC nucleus.

### 2.1.2 Stepwise averaging method

The stepwise averaging method (SAM) is a threshold-based segmentation algorithm. The algorithm works the best if the proportion of non-target areas is larger than the target area. The algorithm takes the average gray value of the image as a threshold. If the gray value of the pixel is smaller than the threshold, it will be unchanged. Otherwise, the gray value of the pixel is set to 0. We can obtain an image after this processing; Secondly, process the image obtained in the previous step in the same way until the mean of the image does not change. Theoretically, the final mean value obtained is the gray value of the background in the bone marrow microscopic image. The background can be removed from the image by using the final mean value. Using this method, we can obtain the gray value of mature red blood cells, the cytoplasm of BMC and the nuclei of BMC in bone marrow microscopic images. The threshold of segmenting the BMC nucleus can be obtained, and the mask image of BMC nucleus can be obtained according to this threshold.

In the practical application, SAM algorithm flow is shown in Algorithm 0.

| Algorithm 0 | |
|---|---|
| **Input:** | An enhanced BMC image using the color transformation, initial parameter $T_i = 0$ and a number of iterations IT=1. |
| **Output:** | A mask image. |
| **S1:** | Calculate the average gray value of the pixels with the gray value greater than $T_i$ in the HSG image, denoted as $T_j$. |
| **S2:** | Calculate the average gray value of the pixels with the gray value which is less than $T_j$ but greater than $T_i$ in the HSG image, denoted as $T_k$. |
| **S3:** | Set $T_i = T_k$ and IT = IT + 1. If IT = 4, stop. Otherwise, repeat Steps 1 and 2. |
| **S4:** | The four average gray values will be obtained for the background, the mature red blood cells, the BMC cytoplasm and the BMC nuclei. Then, use the average gray value of BMC cytoplasm and nuclei for a rough initial segmentation to obtain a mask image. If a pixel value is less than the average value, it is set 0. Otherwise, it will be set to 255. The mask image obtained will contain the BMC nuclei. |

### 2.1.3 Two-stage locating operation

Based on the obtained mask image, we can identify the region of interest (ROI) of the BMC based on the morphological characteristics of the BMC. The ROI is a rectangular area, which contains the BMC nuclei. We propose different algorithms for locating the BMC nuclei and cytoplasm. The specific algorithm for identifying the ROI of the BMC nuclei is shown in Algorithm 1.

| Algorithm 1 | |
|---|---|
| **Input:** | The mask image of BMC nucleus obtained from SAM Algorithm. |
| **Output:** | The ROI of the BMC nuclei. |
| **S1:** | Obtain the outermost contour of the BMC nuclei from the mask image of BMC nucleus. |
| **S2:** | Calculate the circumscribed rectangle of the contours and its center. |
| **S3:** | Determine whether the two neighboring rectangles need to be combined. If two rectangles are overlapped and two centers are falling inside the overlapped area, the two nuclei will be merged into one. |
| **S4:** | The ROI of the BMC nuclei is the circumscribed rectangle of the contours determined by Step 3. |

The specific algorithm for identifying the ROI of the BMC cytoplasm is shown in Algorithm 2.

| Algorithm 2 | |
|---|---|
| **Input:** | The mask image of BMC nucleus obtained from SAM Algorithm. |
| **Output:** | The ROI of the BMC cytoplasm. |
| **S1:** | Calculate the area, S, and the perimeter, L, of the nucleus. |
| **S2:** | Calculate the circularity of the nucleus $CirR = 4\pi S / L^2$. |
| **S3:** | Calculate the reference radius of the nucleus $R = \sqrt{(S/\pi)} = L/(2\pi)$. |
| **S4:** | Calculate the equivalent radius $R_e$ of the circular region of the cell by $$R_e = \begin{cases} 2.6*R, & CirR < T_1 \\ 2.3*R, & T_1 < CirR < T_2 \\ 1.6*R, & T_2 < CirR \end{cases}$$ Where the threshold parameters $T_1$ and $T_2$ are given. According to our experiments, it is suggested to use $T_1=0.46$, $T_2=0.85$. |
| **S5:** | The ROI of the BMC cytoplasm is the circumscribed rectangle of the circular region determined by $R_e$. |

Combining the ROIs of the nucleus and cytoplasm obtained in the two algorithms above by taking the top-left corner and bottom-right corner of two ROIs and re-drawing a new ROI which will include two smaller ROIs, and we can locate the BMC accurately.

### 2.2. Segmentation of BMC

BMC contain both the nucleus and the cytoplasm. To segment BMC from bone marrow samples, first segment the nucleus and then segment BMC according to the distribution of cytoplasm.

### 2.2.1 BMC nucleus segmentation

After the localization of the BMC as outlined in Section 2.1, the nucleus of BMC will be precisely segmented by a threshold segmentation based on the weighted SAM and Otsu's method. The specific segmentation steps are shown in Algorithm 3.

| Algorithm 3 | |
|---|---|
| **Input:** | The ROIs (i.e. a patch) of BMC sample image, mask image and HSG image. |
| **Output:** | The BMC nucleus. |
| **S1:** | Subtract the mask image from the HSG image to obtain a new HSG image (we call the HSG-M image) without the BMC nuclei and Obtain the ROI of the new HSG image based on the same ROI of the BMC. |
| **S2:** | Perform SAM on the ROI of the HSG-M image to obtain the gray values of background, mature red blood cells, cytoplasm and nucleus, which is represented as bT, rT, cT and kT respectively. |
| **S3:** | Perform the Otsu's method on the ROI of HSG-M to obtain the auxiliary threshold wT. |
| **S4:** | Calculate the initial weighted threshold, T, using the thresholds obtained from the SAM and Otsu's method with $T = \gamma * cT + (1-\gamma) * wT$. In our experiments, $\gamma = 0.5$. |
| **S5:** | Based on threshold T, segment the HSG-M to obtain the initial BMC nucleus in the ROI. Then, apply the SAM to the HSG-M and obtained initial BMC nucleus to obtain a more accurate threshold of the BMC nucleus, i.e. $bT', rT', cT'$ and $kT'$. We can then recalculate a new threshold of $T'$ (from Step 4) using the formula $T' = (1-\gamma) * cT' + \gamma * kT'$. |
| **S6:** | Segment the HSG-M to obtain the nucleus image using threshold $T'$. |

Experimental results of algorithm 3 applied to a patch is shown in Fig.3. In this Fig.3, (a) shows the ROI region of BMC; (b) the corresponding color enhanced transformation; (c) the output of Algorithm 3, a BMC nucleus image that may contain impurities; (d) the output of Algorithm 3 is post-processed to obtain an impurity-free BMC nucleus image.

is removed and the BMC cytoplasm is segmented. The steps for the segmentation are shown in Algorithm 4 and 5.

In algorithm 4, the structure of the color weakened transformation image BSG, the BMC cell texture image TeIg, the BMC cytoplasmic white particle image CWPIg and the seed point selection of region growing will be described later in detail.

| Algorithm 4 | |
|---|---|
| **Input:** | The ROIs (i.e. a patch) of BMC sample image; the BMC nucleus obtained from Algorithm 3. |
| **Output:** | The non-BMC mask. |
| **S1:** | The nucleus is removed from the BSG image and the threshold, TB, is obtained using the Otsu's method. If the gray level of a pixel is less than TB, it is assigned into the non-BMC region. |
| **S2:** | Obtain the background and mature red blood cell regions with uniform gray distribution based on the image, TeIg, and assigned into the non-BMC region. |
| **S3:** | According to the number of connected regions in the image CWPIg, it is judged whether the color of the nucleus and cytoplasm of BMC is consistent. If the color is the same, it is considered that the pixel in the image BSG with the gray value equal to zero belongs to the non-BMC region. |
| **S4:** | The boundary point on the circumcircle of the BMC nucleus and the pixel in the non-BMC region determined as the seed point were used to grow the region, and the results were assigned into the non-BMC region. |
| **S5:** | The non-BMC mask is the union of the non-BMC regions obtained in the above four steps. |

In algorithm 5, the structure of the color weakened transformation image BSG will be described later in detail.

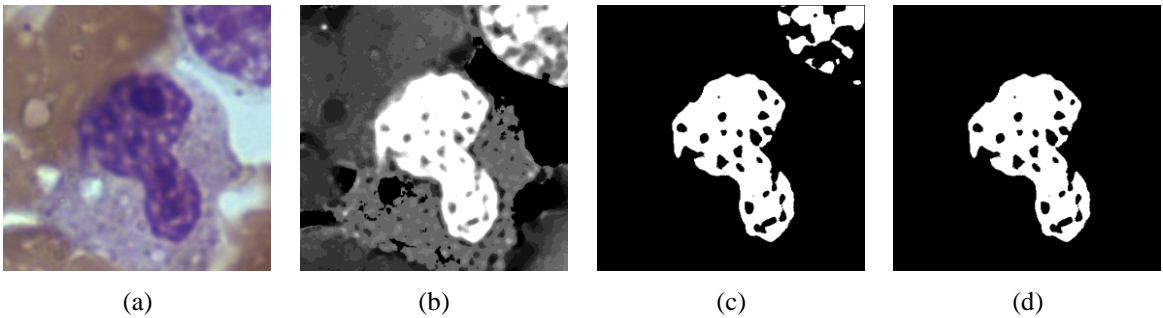

|                (a)                 |                (b)                 |                (c)                 |                (d)                 |

***Fig. 3.*** *(a) the ROI region of BMC; (b) the corresponding color enhanced transformation; (c) the output of Algorithm 3, a BMC nucleus image that may contain impurities; (d) the output of Algorithm 3 is post-processed to obtain an impurity-free BMC nucleus image.*

### 2.2.2 BMC cytoplasm segmentation

Due to the complex background in the bone marrow sample image, we have defined a color weakened transformation to highlight the color characteristics of BMC. This transformation can effectively distinguish the non-BMC region from other regions. Then, the non-BMC region

| Algorithm 5 | |
|---|---|
| **Input:** | The non-BMC mask obtained from Algorithm 4. |
| **Output:** | The BMC mask. |
| **S1:** | In the image BSG, the value of pixels in the non-BMC region are set to 0; Image BSG is then segmented into three categories using the K- |
|      | Means algorithm, the clustered result image, KMImg.                                                                                                                                         |
| S2:  | In the image KMImg, the smallest average gray values are eliminated, and finally the largest connected area is selected as the initial BMC mask image, PdgMask.                             |
| S3:  | According to the pole in the image PdgMask to determine whether the occurrence of adhesion.                                                                                                 |
| S4:  | If the BMC and the surrounding BMC or mature red blood cells were connected, the marker-controlled watershed algorithm will be applied, to obtain more accurate cell boundaries.            |
| S5:  | The BMC mask is the region enclosed by the cell boundaries determined by Step 4.                                                                                                            |

Using algorithm 4 and 5, Fig.4 and 5, respectively, shows the BMC nucleus and cytoplasm of the same and different color, a segmentation result of BMC cytoplasm. In Fig.4, (a) shows the ROI of BMC (large lymphocytes) after BMC localization; (b) shows the region image of the non-BMC; (c) represents the clustered image; (d) represents the average gray value in the clustered image is relatively large two categories as the image PdgMask; (e) represents the maximum connected region in the image PdgMask; (f) represents the pole image; (g) represents the non-BMC marker image; (h) represents the BMC marker image; (i) represents the use of watershed to obtain the image; (j) represents the final BMC boundary image; In Fig.5, (a) shows the ROI of BMC (neutrophilic segmented granulocyte) after BMC localization; The meaning of (b-j) is the same as Fig. 4 (b-j).

The method of generating a transformation image, a texture image, a particle image, and a watershed initial marker will be described in detail below.

(a) Transformation image

It has been observed that there is a great contrast between the foreground and the background in the blue channel of BMC images, and the brightness of BMC is darker than other cells in the green channel of BMC images. In order to make full use of these color features in the algorithm, a new gray-scale mapping of bone marrow microscopic images is obtained by using the color weakening transformation shown in Eq. (2). The Eq. (2) has the same meaning as the Eq. (1).

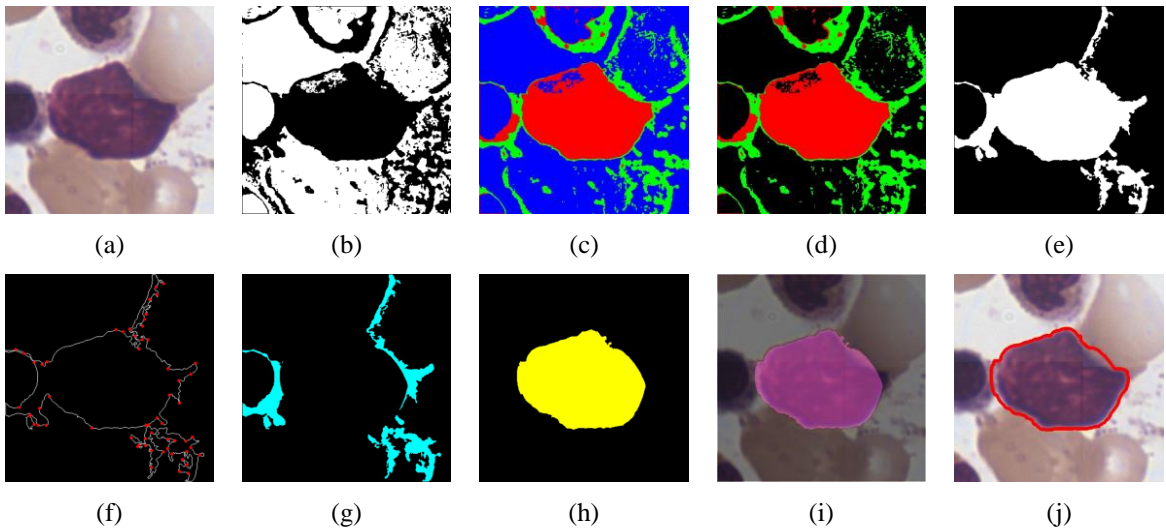

***Fig. 4.*** *(a) the ROI of BMC (large lymphocytes) after BMC localization; (b) the region image of the non-BMC; (c) the clustered image; (d) the average gray value in the clustered image is relatively large two categories as the image PdgMask; (e) the maximum connected region in the image PdgMask; (f) the pole image; (g) the non-BMC marker image; (h) the BMC marker image; (i) the use of watershed to obtain the image; (j) the final BMC boundary image.*

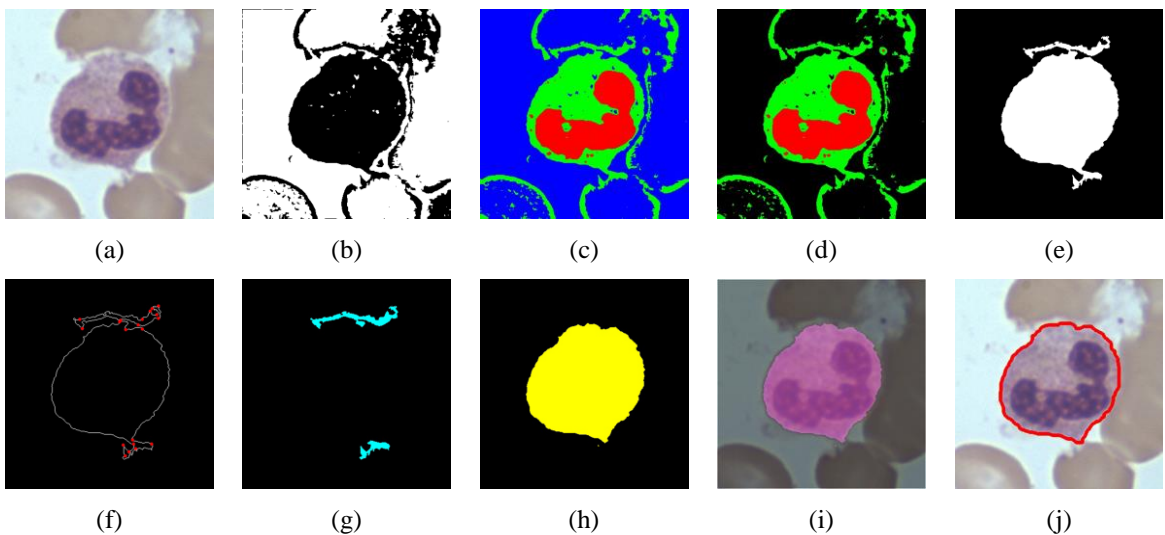

***Fig. 5.*** *(a) The ROI of BMC (neutrophilic split granulocyte) after BMC localization; the meaning of (b-j) is the same as Fig. 4 (b-j).*

$$V(x,y) = \lambda*\{I_B(x,y) - I_G(x,y)\} + (1-\lambda)*\{255 - I_G(x,y)\}$$

$$I_{BSG}(x,y) = \begin{cases} V(x,y), & V(x,y) > 0 \\ 0, & otherwise \end{cases} \quad (2)$$

In equation (2), the values of parameters $\lambda$ can be set according to whether the BMC nucleus and cytoplasm are consistent. If the color is inconsistent, $\lambda=0.5$, otherwise $\lambda=1$. The image obtained by Eq. (2) is called the Blue and Green color weakening image (BSG).

of the current pixel. The formal description is shown in the formula (3).

Where Var(S) represents the variance function of the set S, and w and h represent the length and width of the neighborhood, respectively. Note that in order to increase the speed of acquiring the texture image, the image BSG in the equation (3) is an image after the background is removed. After obtaining the texture image TeIg, if the average gray value of the connected region is zero and its area is greater than a given threshold, it is treated as part of the non-BMC region and recorded as an NWIg image. The image acquisition process is shown in Fig.7, where (a), (b), (c) and (d) correspond to the original ROI image, the image BSG, the image TeIg, and the image NWIg, respectively.

(c) Particle image

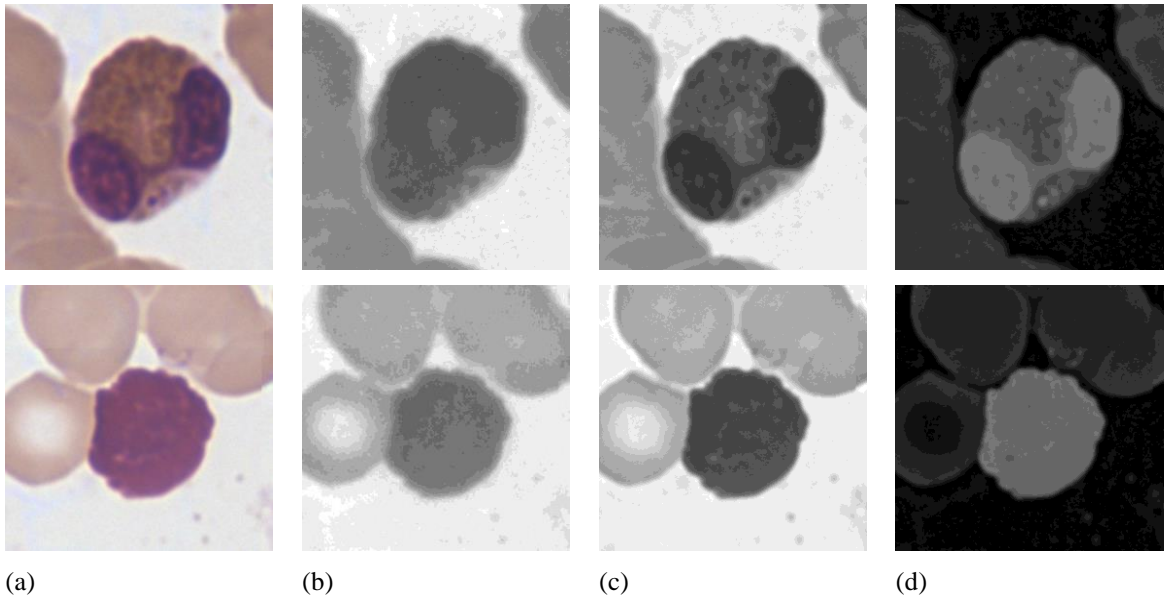

(a)　　　　　　(b)　　　　　　(c)　　　　　　(d)

***Fig. 6.*** *(a), (b), (c) and (d) correspond to the original image, the blue channel image $I_B$, the green channel image $I_G$ and the image BSG, respectively.*

In Fig.6, the first line and second line, respectively, shows the same and different color of the BMC nucleus and cytoplasm, the effect of the color weakened transformation image BSG. (a), (b), (c) and (d) correspond to the original image, the blue channel image $I_B$, the green channel image $I_G$ and the image BSG, respectively.

(b) Texture image

In non-BMC region set, the connected mature red blood cells with BMC are very difficult to remove, but they also have a commonality. It was observed that the distribution of gray values of mature red blood cells was more uniform than that of BMC in most of the bone marrow samples. In order to better remove mature red blood cells, this paper presents the method of generating bone marrow texture images according to the characteristics, and uses gray variance to characterize the texture features of cells. That is, the gray value of the current pixel point is replaced by the variance value of the eight neighborhood gray value

If the color difference of the BMC nucleus and cytoplasm is obvious, there will be white particles on the BMC cytoplasm, which is called the particle image in the BMC cytoplasm. We can estimate whether the color of the BMC nucleus and cytoplasm is consistent according to the number of particles. If the color is consistent, the number of particles contained in the cytoplasm is small, otherwise the number of particles is high.

$$\begin{aligned} LDNA &= \sum_{i=1}^{N} Num(BSGZo_i) \\ Num(ConIg) &= \begin{cases} 1, & ConIg \cap NWIg = \phi \ \& \ A_D > T_s \\ 0, & otherwise \end{cases} \end{aligned} \quad (4)$$

The binarized image BSGZo is obtained by binarizing the image BSG with a threshold of zero. If the connected region in the image BSGZo does not intersect the image NWIg and the connected region area $A_D$ is greater than the set threshold $T_s$, it is treated as part of the cytoplasmic white particle image CWPIg. The number of

$$I_{TeIg}(x,y) = Var\{\ I_{BSG}(x+i, y+j),\ i=-w,...,w,\ j=-h,...,h\ \} \quad (3)$$

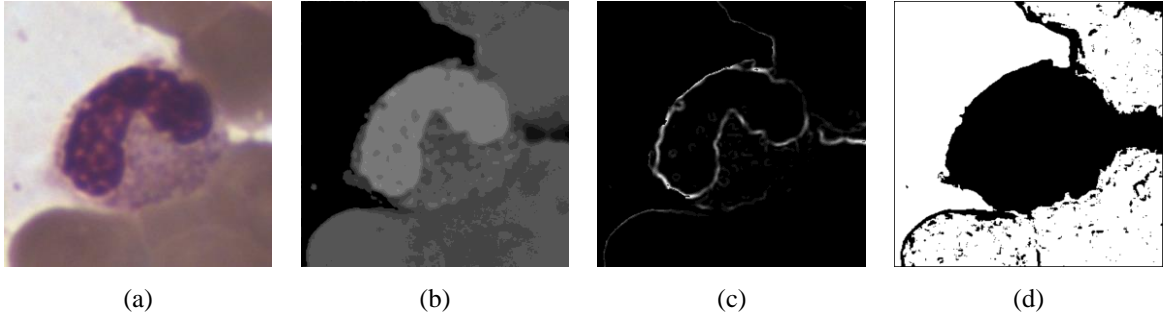

***Fig. 7.*** *(a), (b), (c) and (d) correspond to the original ROI image, the image BSG, the image TeIg, and the image NWIg, respectively.*

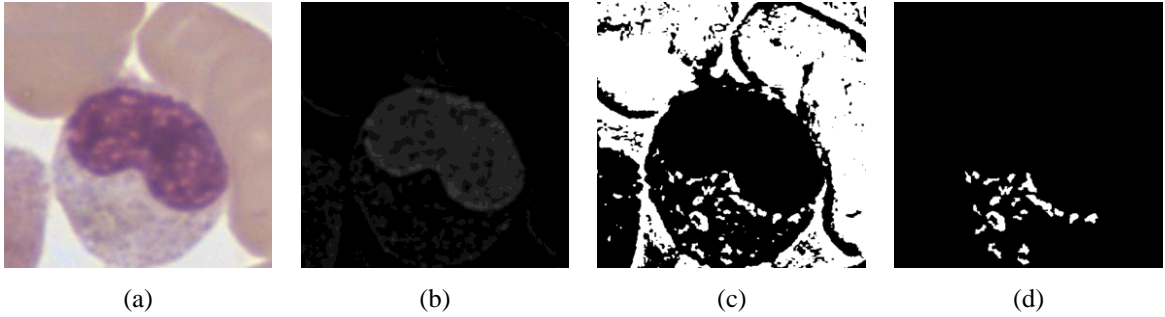

***Fig. 8.*** *(a), (b), (c) and (d) correspond to the original ROI image, the image BSG, the image BSGZo and the image CWIg, respectively.*

particles in the image CWPIg is calculated as (4):

Here, $BSGZo_i$ represents the i-th connected region of the image BSGZo, and Num (ConIg) is a function of determining whether the image ConIg is a white particle of the BMC cytoplasm. If white particle is present, returns 1; otherwise, returns 0.

Fig.8 show the formation of image CWPIg, (a), (b), (c) and (d) correspond to the original ROI image, the image BSG, the image BSGZo and the image CWPIg, respectively.

(d) Watershed initial mark

Whether the current BMC are connected with the surrounding cells is determined by whether the number of poles on the image PdgMask contour is more than two. If more than two, the BMC and other cells have occurred adhesion. In the process of finding the pole, the choice of starting point has a certain impact on the determination of the pole. In order to make generally appear as an increasing trend, the steps of selecting the contour starting point are shown in Algorithm 6.

| Algorithm 6 | |
|---|---|
| **Input:** | The contour of the image PdgMask, the number L of pixels in the cell contour, the center O of the nucleus, and the move step $Sp = 0$. |
| **Output:** | The starting point of the contour. |
| **S1:** | Select the contour point pc with the smallest distance from the center O as the initial contour starting point. |
| **S2:** | Determine whether the connection between the contour point pc and the center O has two intersections with the contour. If so, the pc can be the starting point of the contour and the program is terminated. Otherwise, $Sp = Sp + 1$, the program continues to perform step three. |
| **S3:** | Update the contour point $pc = 2^{Sp} \% L$, where the symbol '%' indicates the modulo operators. |
| **S4:** | Jump to step two. |

In the process of reducing adhesion, if there are multiple adhesions, that is, BMC is connected with multiple mature red blood cells or other non-central BMC, the total number of poles on the contour of BMC will exceed three. At this point, it is necessary to determine that the two poles corresponding to the cells need to be split. However, the actual pole is often not in pairs, there will be some extra pole, this makes it difficult to determine the pole pairs that need to be split, so it is difficult to determine precisely the poles that need to be split at the beginning. In order not to miss pole pair any need to split, this paper uses the method of quickly traversing all possible pole pairs which may need to be split, and then determine the necessity of the division by the increasement of the circularity of the BMC mask after splitting, so as to determine the pole pairs that need to be split.

Before applying the watershed algorithm, first determine the initial mark set of foreground and background objectives, the method is shown as follows. After finding the pole pair that needs to be split, the position distance (Euclidean distance) of the current pole pairs is used as the long axis, and takes a third of its length as a short axis, and then draw a black ellipse in each pair of poles corresponding to the region, and fill it with pixels with a gray scale of zero. The center of the ellipse is the midpoint of each pole pair. Then, the region that is not connected with the central nucleus, which is used as the seed point of the non-BMC, and the remainder region is used as the seed point of the BMC. Finally, the watershed algorithm with marker control is used to obtain the mask image of BMC.

### 2.3. Classification of BMC

The key steps of BMC classifying are extraction and selection [29] of the BMC features. The whole process of BMC classifying is described in detail below.

#### 2.3.1 Feature extraction

In order to classify and count BMC, it is necessary to extract the features of various BMC for training. In this paper, four kinds of BMC features including size, color, texture and morphology were extracted. The meaning of these features will be illustrated below according to the characters of all kinds of BMC.

The shape of cell body and nucleus of metarubricyte is round or oval, and the diameter is 7-10μm. The nucleus is black purple and lump, and its area is less than half of cell body's area. The cytoplasm is carnation or gray red, and has no particles. Based on this, we present several kinds of features, including area ratio of nucleus to cytoplasmic, area ratio of nucleus to cell body, perimeter ratio of nucleus to cytoplasmic, circularity of BMC, the difference of average gray value between the cytoplasm and the nucleus, referred to as Bcnag, and the average gray value of the nucleus, referred to as Bnag in B-component image (i.e., B channel image in RGB color space).

The cell body's shape of mature lymphocyte is round or oval, and the diameter is 12-15μm. Unlike metarubricyte, their nuclei are round or oval, and tend to be on one side. The distribution of nuclear chromatin is uniform and the color is dark purple. The proportion of cytoplasm is relatively large. The cytoplasm is light blue and has a few purple particles. Based on this, we define several features, including perimeter of the nucleus, perimeter of the cell body, Hu moment of the nuclear contour, referred to as Hunc, the average gray value, referred to as Yag, and the ratio of the total number of non-zero pixels to the cytoplasmic area, referred to as Ycr in the Y-component image (i.e., the Y-channel image in the CMYK color space).

The shape of cell body of neutrophilic stab granulocyte is normally round, and the diameter is 10-15μm. However, the diameter of the nucleus is reduced, and the depression exceeds half of the diameter of the nucleus. The shape of the nucleus is similar to "S" shape or "U" shape. Its nuclear chromatin is dense block, and color is deep purple. Cytoplasmic color is pale pink, and it is filled with small and uniform purple particles. Based on this, we define several features, including the number of the nucleus which concave angle is greater than 180 degrees referred as SanV, the area of the nucleus referred as SK, the rectangularity of the nucleus referred as RectR (RectR = SK / the area of the corresponding ROI image), the elongation of the nucleus referred as KExtR (KExtR = min (w, h) / max (w, h), w and h represent the width and height of the nucleus external rectangle respectively), the length of the nucleus centreline after skeleton thinning referred as Lk, and the thinning degree of the skeleton thinning algorithm, referred to as RL (RL = LK / SK).

The shape of neutrophilic split granulocyte's cell body is generally round, and the diameter is 10-13μm. The nucleus is lobulated, and most has 2-3 leaves. The leaves are connected by filaments. However, the nucleus is sometimes overlapped, resulting in hidden of nuclear filament. These cells are not easily distinguished from neutrophilic stab granulocytes. Its chromatin is the same as neutrophilic stab granulocyte, and its cytoplasm is pale pink and is filled with neutral particles. Based on this, we define several features, including erosion times when two or more contours appear for the first time, referred as ErTwo, erosion times when the nucleus area becomes zero, referred as ErZero, the adhesion degree of the nucleus, referred as KAd (KAd = ErTwo / ErZero), the number of lobes in the nucleus, referred as NlV, the area ratio of the nucleus inner contour to the whole contour, referred as CtrV.

Other cells have their own characteristics. The eosinophils' cell body is round, and its diameter is 10-16μm. Its nucleus is similar to the nucleus of neutrophils, and its cytoplasm is filled with golden particles. The basophils' cell body is round, and its diameter is 10-12μm. Its nucleus is blurred rod or lobulated, and its cytoplasm and nucleus have little dark purple particles. The monocyte cell's shape is round or irregular, and the diameter is 12-20μm. The color of cytoplasm is gray-blue or gray-coloured. It's translucent, just like ground-glass. There are some tissue cells and plasma cells, etc. All in all, there are many types of cells and their morphologies are different in this category. Based on this, we define some features, including the fractal dimension information of S component (S channel image of HSV color space), referred as Sfdi, three optical densities of the nucleus -- ND1, ND2 and ND3, and three gray-level co-occurrence matrices extracted according to the R component of the nucleus (i.e. the R - channel image in the RGB color space), referred as RNglcr. Similarly, the three optical density characteristics CD1, CD2 and CD3 of the cytoplasm were extracted, and the three gray-level co-occurrence matrices were extracted according to the Y component of the cytoplasm, referred to as YCglcm.

**Table 1** Defined 39 features. Note that, "B" denotes the cell body of BMC; "N" denotes the nucleus of BMC; "C" denotes the cytoplasm of BMC.

| Size features | N's Area | N's perimeter | N's KExtR | N's RectR | N and C area ratio | N and B perimeter ratio | B's perimeter | C's Area | Ycr | CtrV |
|---|---|---|---|---|---|---|---|---|---|---|
| Color features | | Yag | | | Bnag | | | Bcnag | | |
| Texture features | | Sfdi | | | RNglcr | | | YCglcm | | |
| morphological features | NlV | Hunc | N' circularity | B' circularity | ND1, ND2 and ND3 | CD1, CD2 and CD3 | Connected Region number | | Eccentricity | |
| Customized morphological | | ErTwo | | KAd | | SanV | | Lk | | RL |

In order to classify BMC accurately, 39 different features were selected according to the characters of all kinds of BMC, as shown in Table 1.

### 2.3.2 Classification method

In normal circumstances, the human bone marrow usually contains a large number of granulocytes, red blood cells, lymphocytes, monocytes and a small amount of plasma cells, tissue cells and so on. Their proportion in the bone marrow is shown in Table 2. Note that only the proportion of more than %1 is listed in Table 2.

lymphocytes, which naive lymphocytes can be ignored. The remaining monocytes and plasmacytes accounted for no more than 3%, which will be divided into other cells, when classified. But for BMC from different periods within each class, such as the original red blood cells, early red blood cells and young red blood cells, it is difficult to find the features of the class, that is, it has no obvious statistical significance, so it is difficult to classify them. Therefore, only five types of cells with statistical significance and large proportion were selected for classifying. These five types of cells were: neutrophilic split granulocyte, neutrophilic stab granulocyte, metarubricyte, mature lymphocytes and other cells (eosinophils, basophils, monocytes, etc.).

In this paper, according to the features extracted in Section 2.3.1, using support vector machine (SVM) [30-32], random forest (RF) [33], artificial neural network (ANN) [34] and other machine learning methods for classification training. Experiments show that support vector machines have the best experimental results, followed by artificial neural networks.

### 3. Results and discussion

**Table 2** The proportion of various types of BMC

| Cell name | | | Normal range |
|---|---|---|---|
| | | promyelocyte | 0.710 - 1.930 |
| granulocytic system | neutrophil | myelocyte | 5.690 - 10.71 |
| | | metamyelocyte | 9.290 - 16.47 |
| | | stab nucleus | **12.39 - 20.33** |
| | | split nucleus | **10.76 - 20.10** |
| | eosinocyte | metamyelocyte | 0.250 - 1.370 |
| | | split nucleus | 0.260 - 2.360 |
| erythrocyte system | | prorubricyte | 0.240 - 1.340 |
| | | polychromatic erythroblast | 5.910 - 10.81 |
| | | metarubricyte | **6.960 - 14.02** |
| lymphocyte | | mature lymphocytes | **14.39 - 26.25** |
| monocyte | | mature monocytes | 0.250 - 2.030 |

In abnormal circumstances, the human bone marrow may also contain basophilic stab granulocyte, early erythrocytes, juvenile erythrocytes, late giant erythrocytes, primitive lymphocytes, shaped lymphocytes, primordial monocytes, naive monocytes, primitive plasma cells, juvenile plasma cells, tissue eosinophil, abnormal tissue cells, phagocytic cells, mast cells, hair cells, immature cells, lymphoma cells, metastatic cells and other cells.

As can be seen from Table 2, the proportion of neutrophilic stab granulocyte and neutrophilic split granulocyte exceeds 50% in the granulocyte system, metarubricyte in the red blood cell system accounted for more than 50%. In addition to granulocyte system and erythrocyte system, the largest proportion is the

The sample set used in this study is from Wuhan Landing Medical High Technology Co., Ltd., a total of 48037 100-times microscopic images of bone marrow samples, which contain a variety of different backgrounds, including almost all members of the BMC family, young to mature granulocytes, lymphocytes, and monocytes. Each bone marrow sample was obtained by using DFK23G274 color camera (volume 29*29*57m3, pixel size4.4um*4.4um) on the Olympus microscope BX41 (numerical aperture of 1.30). In the experiment, we divided it into two parts: training set and testing set. The training set consists of 40033 samples including neutrophilic split granulocyte, neutrophilic stab granulocyte, metarubricyte, mature lymphocytes and the outlier (all other cells not listed), with the corresponding samples 2747, 10129, 9890, 5724 and 11543, respectively. The testing set contains 8004 samples, including neutrophilic split granulocyte, neutrophilic stab granulocyte, metarubricyte, mature lymphocytes and the outlier with corresponding samples, 549, 2025, 1978, 1144 and 2308, respectively.

In this paper, we compare the SVM-based method [26], RF-based method [25], ANN-based method [23], Adaboost-based method [22] and Nbayes-based method [24]. Table 3 shows the recognition rates for the various classifiers.

**Table 3** Shows the recognition rates for the various classifiers. Note that, NSTG denotes neutrophilic split granulocyte, NSBG denotes neutrophilic stab granulocyte, MBE denotes metarubricyte, MLS denotes mature lymphocytes, OCS denotes other Cells and ARR denotes average recognition rate.

|  | NSTG | NSBG | MBE | MLS | OCS | ARR |
|---|---|---|---|---|---|---|
| **SVM-based method [26]** | **87.43%** | **87.06%** | **81.24%** | **82.52%** | **99.22%** | **87.49%** |
| RF-based method [25] | 33.15% | 83.31% | 87.66% | 49.21% | 74.00% | 65.47% |
| ANN-based method [23] | 51.91% | 81.73% | 83.67% | 62.85% | 73.31% | 70.69% |
| Adaboost-based method [22] | 46.81% | 75.51% | 82.05% | 50.44% | 71.75% | 65.31% |
| Nbayes-based method [24] | 63.21% | 51.85% | 37.16% | 81.56% | 34.23% | 53.60% |

**Table 4** SVM related parameters

| svm_type | kernel_type | degree | gamma | coef0 | C | nu | p | max_iter |
|---|---|---|---|---|---|---|---|---|
| C_SVC | RBF | 10.0 | 0.09 | 1.0 | 10.0 | 0.5 | 1.0 | 1000 |

It can be concluded that the SVM classifier's classification performance is significantly better than the other four classifiers from Table 3. Among them, the classification accuracy of neutrophilic split granulocyte and other cells is much better than other classifiers, and their classification accuracy is 87.43% and 99.22% respectively. Therefore, we finally adopted the SVM classifier for bone marrow cells detection. In the experiment, the relevant parameters of SVM are shown in Table 4.

## 4. Conclusion

Based on the transformed images, the impurities in the bone marrow microscopic image are removed by region growing and other method, the candidate mask was selected by K-Means clustering, and the connected cells were split by the watershed algorithm. Therefore, a fast and accurate BMC segmentation algorithm was proposed. Then, machine learning was used to classify and count the BMC, and compared with support vector machine, random forest, artificial neural network, enhanced learning, and Bayesian classifier.

In the future work, we will make further improvement for the case of cell adhesion, such as the gradient image obtained by Sobel operator and canny operator for obtaining the edges on the gradient image. If the missing contour exists, theoretically we can find a more natural BMC boundary. Then, under the precondition of obtaining more samples, we try to use the supervised machine learning method to segment the complex BMC to improve the robustness of the algorithm. If the mass sample can be obtained, we will try to use the deep learning methods for classification and identification.

## 5. Acknowledgements

This work was supported by the National Natural Science Foundation of China (Grant Nos. 61370179 and Grant Nos. 61370181). Author would like to acknowledge Zhongnan Hospital of Wuhan University and Wuhan Landing Medical High-Tech Company, limited for providing the test samples. In particular, the author would like to thank the reviewers for their constructive comments in improving the quality of the manuscript.